\documentclass{article}
\usepackage{arxiv}
\usepackage{amsmath,amssymb,amsfonts}
\usepackage{graphicx}
\usepackage{textcomp}
\usepackage{bmpsize}
\usepackage{xcolor}
\usepackage{lipsum}
\usepackage[colorlinks=true,urlcolor=black]{hyperref}
\def\BibTeX{{\rm B\kern-.05em{\sc i\kern-.025em b}\kern-.08em
    T\kern-.1667em\lower.7ex\hbox{E}\kern-.125emX}}
\usepackage{hyperref}

\usepackage{
    bm,
    amsmath,
    amssymb,
    booktabs,
    footnote,
    graphicx,
    hyphenat,
    multirow,
    pgfplots,
    url,
    hyperref,
    siunitx,
    nicefrac,
    pifont,
    threeparttable,
    cleveref,
    adjustbox,
    enumitem,
    todonotes,
    algorithm,
    algpseudocode
}
\usepackage[numbers]{natbib}
\title{Blackbox Attacks on Reinforcement Learning Agents \\Using Approximated Temporal Information}

\makeatletter
\newcommand{\printfnsymbol}[1]{%
  \textsuperscript{\@fnsymbol{#1}}%
}
\makeatother

\newcommand{\wbox}{White-box}
\newcommand{\gbox}{Grey-box}
\newcommand{\bbox}{Black-box}

\newcommand{\eg}{\textit{e.g\@.}}

\newcommand{\etal}{\textit{et~al\@.}}

\newcommand{\cmark}{\ding{51}}%
\newcommand{\xmark}{\ding{55}}%

\author{%
Yiren Zhao\thanks{
    Ilia Shumailov, Han Cui and Yiren Zhao
    contributed equally to this work.
}\\ University of Cambridge \And%
Ilia Shumailov\footnotemark[1]\hspace{4pt}\\ University of Cambridge \And%
Han Cui\footnotemark[1]\hspace{4pt}\\ University of Bristol \And%
Xitong Gao\\ SIAT\And%
Robert Mullins\\University of Cambridge \And%
Ross Anderson\\University of Cambridge
}

\begin{document}

\maketitle

\begin{abstract}
Recent research on reinforcement learning (RL) has suggested that trained agents are vulnerable to maliciously crafted adversarial samples.
In this work, we show how such samples can be generalised
from \wbox~and \gbox~attacks to a strong \bbox~case, where the attacker has no knowledge of the agents, their training parameters and their training methods.
We use sequence-to-sequence models to predict a single action or a sequence of future actions that a trained agent will make.

First, we show our approximation model, based on time-series information from the agent, consistently predicts RL agents' future actions with high accuracy in a \bbox~setup on a wide range of games and RL algorithms.

Second, we find that although adversarial samples are transferable from the target model to our RL agents,
they often outperform random Gaussian noise only marginally.
This highlights a serious methodological deficiency in previous work on such agents; random jamming should have been taken as the baseline for evaluation.

Third, we propose a novel use for adversarial samples
in \bbox~attacks of RL agents: they can be used to trigger a trained agent to misbehave after a specific time delay. This appears to be a genuinely new type of attack.
It potentially enables an attacker to use devices controlled by RL agents as time bombs.
\end{abstract}

\section{Introduction}
Deep neural networks (DNNs) have good performance on a wide spectrum of tasks, ranging from image classification \cite{krizhevsky2012imagenet},
object detection \cite{ren2017faster} and
emotion recognition \cite{nicholson2000emotion}
to language processing \cite{vaswani2017attention}.
Recent advances in reinforcement learning (RL) demonstrate that DNNs can learn policies that solve complex problems by mapping raw environment inputs directly to an action space.
Trained deep RL agents show human-level or even superhuman performance in playing Go \cite{silver2016mastering} and Atari games \cite{mnih2013playing}.
Following this success, RL agents are starting to be exploited in safety-critical fields, such as robotics \cite{levine2016end}, as well as in recommendation systems \cite{zheng2018drn} and trading \cite{nevmyvaka2006reinforcement}.

Researchers have also found that DNNs are vulnerable to crafted adversarial perturbations.
DNN-based image classifiers can produce incorrect results
when inputs are injected
with small perturbations that are not perceptible by humans \cite{goodfellow2014explaining}.
Attackers can thus create adversarial examples that
cause DNN-based systems to misbehave, including systems for
face recognition \cite{kurakin2017adversarial}
and autonomous driving \cite{eykholt2018robust}.
One feature of adversarial samples is their transferability.
Adversarial inputs that affect one model often affect others too \cite{papernot2016transferability,zhao2018understanding}.
Therefore, adversarial samples can easily affect machine learning (ML) systems at scale.

However, attacking RL agents is different from fooling image classifiers. First, there is no notion of supervised labels in an RL agent, as its performance is assessed purely on the rewards it earns in an episode of game playing.
Second, an action that a well-trained RL agent performs depends on a sequence of observations containing historical data that an attacker cannot modify.
Finally, a single misprediction does not usually cause serious disruption to the agent.
These properties make it a challenging problem to
create realistic \bbox~adversarial attacks on RL.
We will return to the discussion of these properties in~\Cref{sec:motivation}.

Researchers have recently looked for effective attacks
on RL agents \cite{huang2017adversarial,lin2017tactics}
in \wbox~or \gbox~setups by assuming the attackers have access
to some of the agent's internal states or training methods.
These attacks aim to force an agent to take unwanted
actions so that its game score is decreased.
In a typical \gbox~setup, the attackers can access partial information
of the target agent or its training environment, and retrain another agent to approximate the target agent \cite{huang2017adversarial, lin2017tactics}.

Here we go much further. We make a strong \bbox~assumption that attackers have no knowledge of either the agent's parameters or its training methods.
We cannot therefore just retrain another RL agent to do the the same task.
We rather study how approximating RL agents can be formed as a sequence-to-sequence learning problem, as in imitation learning.
RL agents learn over time using a sequence of observations, and produce actions that can also be treated as a temporal sequence.
The question of approximating RL agents can be treated
as a sequence-to-sequence (seq2seq) learning problem
which is familiar from language-translation tasks~\cite{sutskever2014sequence,gehring2017convolutional}.
Given a sequence of observations of the target agent,
we build a seq2seq model to predict its future actions from watching how the target performs, with no knowledge of its internals and training methods.
We show empirically that the model predicts future sequences of actions
consistently with over $80\%$ accuracy on three different games with three different RL training algorithms.
The action sequence prediction we generate is called the \textit{approximated temporal information}.

We then demonstrate how to use the seq2seq model to produce adversarial samples.
In prior work under \wbox~or \gbox~assumptions, researchers exhibited adversarial samples that could be used to reduce the target agent's game score by feeding them in as perturbations to its game input. In our experiments, we show that our \bbox~attack can decrease the target agent's game score.

One interesting thing that we discovered is that the results we get, and also the results achieved by previous researchers, only marginally outperform random Gaussian noise. In other words, their attack was weak; we can do as well, with significantly fewer assumptions.

The second interesting thing we found is that our model is useful in a novel attack.
By an appropriate perturbation of the current input, we can influence a future action after a specific delay with a high probability of success. This gives us a novel {\em time-bomb} attack on RL agents.

The contributions of this paper are the following.
\begin{itemize}
	\item We provide an open-source framework to perform \bbox~attacks on RL agents.
	\item We show that, for the first time, attacking RL agents is possible in strong \bbox~assumption without retraining another RL agent to perform the same task.
	\item We build sequence-to-sequence models to predict
	future agent actions with above $80\%$ accuracy on
	Cartpole and Atari games trained with different RL algorithms.
    \item We demonstrate that although adversarial attacks cause trained agents to perform unwanted actions more frequently, random Gaussian noise can still be as effective as adversarial attacks if the goal is simply to reduce the game score of a target agent. When evaluating attacks on such agents, random noise jamming should be the baseline case.
	\item We show a novel time-series \bbox~attack, the time-bomb attack, that uses adversarial samples to flip actions after a specific delay. This attack opens up a new frontier in adversarial RL.
\end{itemize}

\section{Motivation}
\label{sec:motivation}

Recent advancements in deep learning have led to the adoption of DNN-based control mechanisms for a range of tasks.
Attacks against DNNs have become more and more sophisticated, yet in practice very few of those attacks are an actual threat to a deployed system.
It is common enough for researchers to focus on the interesting attacks in a new technical field, and find the practical subset of attacks only with time.
In the case of adversarial ML, the interesting attacks are often considered to be those that produce perturbations which are imperceptible to humans.
Second, almost all of them focus on disrupting a single standalone task.

In the real world, however, DNNs are usually a component of a larger stateful system with both space and time aspects.
While the space aspect has attracted some research~\cite{engstrom2019exploring}, the timing side has been much less explored.
This work aims at finding practical attacks on systems that control some critical resource for a period of time, and on observing their behaviour closely so as to find just the right time to attack.

RL is used to train agents that interact continuously with the world in order to solve a particular problem.
The agent's performance fully depends on all of the decisions it made, and can be measured directly using the objective function it aims to optimise.
We assume that we are only passively observing.

As an example, consider air combat where the opponent's aircraft have been observed performing a set of manoeuvres to follow and intercept targets.
Our goal is to learn from observational data and make as good a model as we can of the agent flying the plane. We then examine this to find the best possible evasive manoeuvres. Our baseline is random evasion; we want to know whether we can devise a better set of tactics.

Our threat model covers a much larger number of different use-cases, so we evaluate our attack using different RL algorithms with different objective functions.
Although the evaluation can not represent all of the use-cases, it at least shows that in practice it is possible to perform attacks in a fully~\bbox~fashion. Finally we show that the work generalises in that we can learn to model an agent with an unknown objective function and thus make predictions about its future behaviour. So long as we can observe the agent long enough, we may be able to predict its behaviour enough to disrupt it, and in some cases even to cause disruption after a known delay.


\begin{figure*}[!th]
	\centering
    \includegraphics[width=\textwidth]{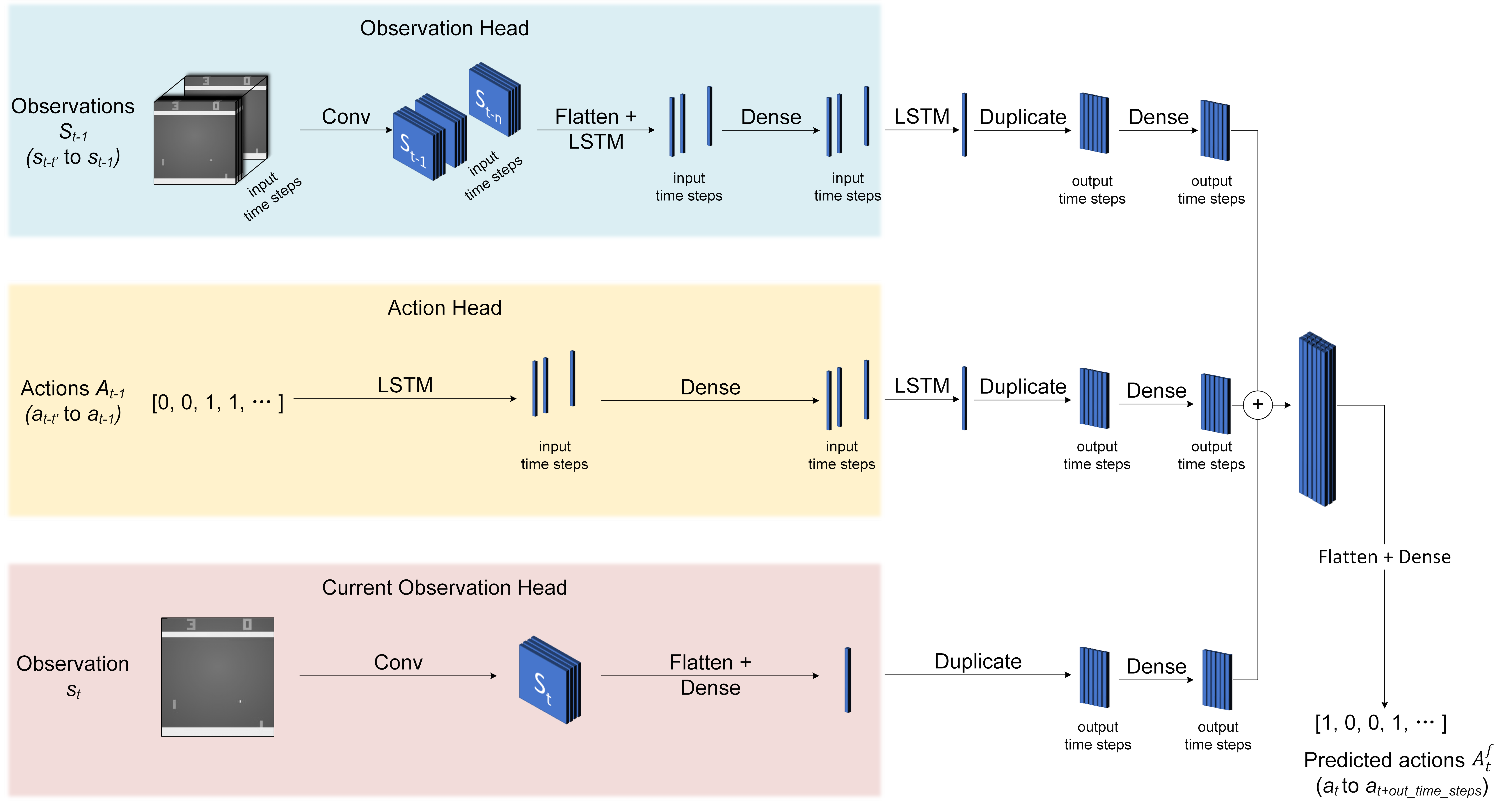}
    \caption{
        An illustration of the sequence-to-sequence network's architecture,
        the network is multi-head: observation head ($S_{t-1}$), action head ($A_{t-1}$) and
        current observation head ($s_t$).
        The output of the seq2seq model is a sequence of predicted future actions ($A^{f}_t$).
        The details of designing each head
        is game-dependent and shown in \Cref{tab:approximator}.}
    \label{fig:network}
\end{figure*}

\section{Related Work}
\label{sec:related}

Deep neural networks (DNNs) help RL to scale to complex decision problems.
A large state or action space make learning intractable for
traditional Markov Chain Monte Carlo methods.
\citeauthor{mnih2015human} firstly proposed
combining deep convolutional neural networks with reinforcement learning (DQN).
DQN is a value-based approximation, where we use a neural network to approximate the action probabilities
at a given state~\cite{mnih2013playing,mnih2015human}.
\citeauthor{mnih2015human} demonstrated that DQN achieves super-human performance
on a series of Atari games.
Later on, \citeauthor{mnih2016asynchronous} proposed the
asynchronous advantage actor-critic method, where the actor performs
actions based on its underlying DNN, a critic scores the performance
of the actor, and the actor then updates its DNN parameters based on the
score received~\cite{mnih2016asynchronous}.
The Actor-critic method can be viewed as a combination of value-based and policy-based methods since the actor is learning a policy function while the critic is learning a value function.
\citeauthor{hessel2018rainbow} built on top of the DQN framework and combined it with a range of possible extensions~\cite{hessel2018rainbow}, including double Q-learning \cite{van2016deep}, prioritised experience replay \cite{schaul2015prioritized},
and dueling networks \cite{wang2015dueling}.
They then demonstrated empirically that their algorithm Rainbow outperforms DQN and Actor-critic on a range of tasks.

\citeauthor{goodfellow2014explaining} proposed
the Fast Gradient Sign Method (FGSM) to produce adversarial samples using DNN gradients~\cite{goodfellow2014explaining}.
The samples contain small perturbations that are imperceptible by humans, yet DNNs produce high-confidence incorrect answers on these inputs. They further demonstrated that attackers can create adversarial inputs targeting particular labels.
Later researchers showed how to apply scaled gradients iteratively to the original input image
\cite{kurakin2017adversarial,madry2018towards}.
Iterative methods such as the projected gradient descent (PGD) attack, proposed by \citeauthor{madry2018towards},
show better performance than single step attacks (\eg~FGSM)~\cite{madry2018towards}.
The Carlini \& Wagner attack (CW) teaches how
to generate adversarial samples by solving an optimisation problem efficiently~\cite{carlini2017towards}.
However, the large number of iterations required makes it expensive to execute in real time.
Adversarial attacks are not limited to a \wbox~setup.
Assuming that the attackers have only the abilities of querying a classifier, they can efficiently estimate the gradients of the hidden classifier and thus build adversarial inputs based on the estimated gradients~\cite{ilyas2018blackbox, brendel2017decision}.

A major threat from adversarial samples is their transferability. This refers to the fact that an adversarial samples may cause misclassifications on different classifiers.
\citeauthor{szegedy2013intriguing} observed
that models of different configurations can easily misclassify on the same set of adversarial inputs~\cite{szegedy2013intriguing}.
\citeauthor{zhao2018understanding} later pointed out that
transferability is also found on
a range of pruned or quantised networks~\cite{zhao2018understanding}.

\citeauthor{huang2017adversarial}
were among the first to apply adversarial attacks to RL agents~\cite{huang2017adversarial}.
They evaluated FGSM attacks in both \wbox~and \bbox~settings, finding them effective on RL agents, while various RL training algorithms show different levels of robustness.
However, they assumed attackers had access to the agent's training environments and DNN structures.
\citeauthor{pattanaik2017} further extended this approach and constructed a loss function that
can be used to attack RL systems more effectively~\cite{pattanaik2017}.
They also introduced adversarial training for RL agents and evaluated its performance.

\citeauthor{behzadan2017vulnerability} took a similar approach
to \citeauthor{huang2017adversarial}; they designed a replica of the target agent and conducted FGSM and JSMA attacks on it to create transferable adversarial samples~\cite{behzadan2017vulnerability}.
In their evaluation, they allowed attackers to periodically sample both the parameters and network architecture of the target agent \cite{behzadan2017vulnerability}.
They further assumed the attacker has full access to the training environment and training hyperparameters.

In our work, we assume a strong \bbox~setup, where attackers have no knowledge of the training environments, methods or parameters.
In the observation phase, they can only observe how an agent interacts with the game. In the attack phase, they can apply specific perturbations to the target environment.



\citeauthor{lin2017tactics} argue that the attack presented by \citeauthor{huang2017adversarial} contains an impractical assumption, namely that attacks are conducted at every step; they also noted that the attacks ignored the facts that the subsequent observations are correlated, and one should not use adversarial samples every step because of the detection risk.
\citeauthor{lin2017tactics} reported a new timed-strategic attack: an adversary can attack a quarter as often, but achieve the same reward degradation as if the attack was performed at every step.
They also explained that the attacks against RL are different from those on image classifiers: first, the goal is not to cause misclassification, but reward reduction; and second, future observations depend on the current observation.
In order to forecast the target agent's future actions, they exploited a video-sequence prediction model for sequence planning.
However, their generation of adversarial samples still requires access to the target agents' parameters.
A more detailed comparison is presented in~\Cref{sec:threat}.

In this paper, we approximate the behaviour of RL agents and reduce the approximation to a classification problem using sequence-to-sequence models.
The approximation of an RL agent can be seen as an imitation learning or apprenticeship learning problem that has been thoroughly studied in the field of ML.
The imitation learning field tries to learn the policy of an expert, given inputs as a sequence of observation-action pairs. The big challenge in the field of imitation learning is handling cases where expert annotations are not available. The approaches can largely be separated into two: active~\cite{torabi2018behavioral,judah2012active} and passive~\cite{osa2018algorithmic}. Active imitation learning aims to fill the gaps by asking the expert to annotate additional information, whilst passive assumes that no additional data can be collected. In this work we assume passive imitation learning as we cannot query the target agent in a~\bbox-setup.

Finally, we are not the first to follow the classification
approach in imitation learning.
\citeauthor{syed2010reduction} reduced apprenticeship learning to classification~\cite{syed2010reduction}. 
But, to the best of our knowledge, we are the first to perform multi-step prediction in a passive imitation learning setting.
Typical passive imitation learning setup only requires to plan an action given the current state. This is similar to our sequence-to-sequence approximation with a single action output. That approach, however, does not work for predicting multiple future actions because of the~\bbox~assumptions~\cite{osa2018algorithmic}.

\section{Method}
\subsection{Preliminaries}
At each time period $t$,
the environment provides a state $s_t$ to an agent, whether an image, or scalar inputs from sensors.
The agent responds with an action $a_t$, and the environment
feeds back a reward $r_t$.
The interaction between the agent and the environment forms a Markov Decision Process, and
an agent learns a policy $\pi$ that describes a probability
distribution on the action space given the current state $s_t$
\cite{mnih2013playing,mnih2016asynchronous,hessel2018rainbow}.
The policy $\pi$ is trained to maximise the
expected discounted return $R_e$,
where $R_e = \sum_{i=0}^{l} e^{\lambda i} r_{t+i}$,
$l$ is the total number of steps in an episode of game play
and $\lambda$ is the discount factor.
A trained agent typically takes states or state-action pairs as inputs
to decide what action to take ($a_{t}$).
Here, we provide a relaxed notion to the inputs of the agent's policy, since
popular techniques such as frame stacking can provide an agent with a history
of its inputs.
Let us consider the sequence of states
$S_t = ( s_t, s_{t-1}, \cdots, s_{t-n} )$
and the sequence of actions $A_t = ( a_{t}, a_{t-1}, \cdots, a_{t-n} )$,
where $n$ is the length of the sequences and $ 0 \leq n \leq t$;
we have the policy function:
\begin{equation}
    a_{t} = \pi(S_t, A_{t-1}).
\end{equation}

To mount an adversarial attack on a trained policy $\pi$,
we construct an adversarial sample for the current state $s_t$ only,
as previous states in $S_t$ are historical and cannot be modified.
The objective is to compute
a small perturbation $\delta_t$
for the input to the policy function,
which now becomes
$\hat{S}_t = ( s_{t} + \delta_t, s_{t-1}, \cdots, s_{t-n} )$,
such that the resulting action
\begin{equation}
    \hat{a}_{t} = \pi(\hat{S}_t, A_{t-1})
\end{equation}
differs from the intended action $a_{t}$,
in such a way as to give a successful attack.
It should be noted that an MDP setup is assumed for simplicity of explanation.
In practice, the problem description often
changes from a complete MDP to a Partially Observed
MDP (POMDP) problem.
This implies that the agent does not have the knowledge of the entire
environment, only a subset.
RL agents usually assume a POMDP environment.

\subsection{Threat Model}
\label{sec:threat}
We assume that the attacker can modify the environment so that the target agent
receives a perturbed state $\hat{s_t}$ and  aims
at changing a future action $a_{t+m}$ to $\hat a_{t+m}$.
We assume that past states
and target agent memory cannot be modified.
If frame stacking is done on the agent side, the attack should not
change previously stacked states.
Since the attack is \bbox, we assume that:
\begin{enumerate}
    \item The attacker has no knowledge of the target agent, its training method or the training hyperparameters;
    \item The attacker has no access to the training environment.
\end{enumerate}
\begin{table*}[!h]
    \caption{
        A comparison to show our attack holds a stronger Black-box assumption.
        The first row shows whether the attacker has access to the agent's
        weights, the agent's DNN structures, the training algorithm
        and the training environment. Huang \etal~
        proposed two Black-box setups and we enumerated them as 1 and 2.
        }
    \footnotesize
    \centering
    \adjustbox{max width=\linewidth}{%
    \begin{tabular}{llllll}
    \toprule
    Attacker Access      & DNN weights   &DNN structure   & Train algorithm   & Train environment \\
    \midrule
    \citeauthor{huang2017adversarial} 1
                         & \xmark        & \cmark              & \cmark       & \cmark              \\
    \citeauthor{huang2017adversarial} 2
                         & \xmark        & \cmark              & \xmark       & \cmark              \\
    \citeauthor{behzadan2017vulnerability}
                         & \xmark        & \xmark              & \cmark       & \cmark              \\
    \citeauthor{lin2017tactics}
                         & \cmark        & \cmark              & \xmark       & \xmark              \\
    Ours
                         & \xmark        & \xmark              & \xmark       & \xmark              \\
    \bottomrule
    \end{tabular}
    \label{tab:compare}
    }
\end{table*}

These assumptions give a very strong \bbox~box setup
as illustrated in \Cref{tab:compare}.
We summarise the differences from previous work in~\Cref{tab:compare}.

\citeauthor{huang2017adversarial} present two different cases:
one assumes no access to the internal DNN weights of the agent, while the
other one further constrains access to the RL training algorithm~\cite{huang2017adversarial}.
\citeauthor{behzadan2017vulnerability} assume the DNN structure is
not known to the attacker~\cite{behzadan2017vulnerability}.
Finally, \citeauthor{lin2017tactics} assume that the attackers have no
knowledge of the training methodology~\cite{lin2017tactics}.

The only explicit agent assumption we make in this work is that the
episodes were collected from the agent and the agent is in evaluation mode i.e. random
exploration is turned off and no more training is done.
In all previous work the approximation of an agent involves retraining 
another RL agent to perform the same task as the target agent.
Adversarial samples are then generated from the clone agent.
In this paper, however,
we cannot just retrain because of the strong \bbox~setup.

\begin{table*}[h]
    \caption{Black-box approximation network configurations and accuracies for different games.
        We measure how accurate our network predicts the next action or a sequence
        of actions with respect to a RL agent trained with DQN.
        \textbf{Seq} means the approximation network predicts the next
        10 consecutive actions from time $t$ ($a_{t}$ to $a_{t+9}$),
        if it is not \textbf{Seq}, only a single action ($a_{t}$) is predicted.
        \textbf{Obs Head}, \textbf{Action Head} and \textbf{Current Obs Head}
        refers to different parts of the networks as illustrated in \Cref{fig:network}.
        \textbf{Input Seq} shows the input sequence length for both
        actions and observations.
        }
    \footnotesize
    \centering
    \begin{tabular}{lllllll}
    \toprule
    \textbf{Game}        & \textbf{Acc}  & \textbf{Obs Head} &\textbf{Action Head}  &\textbf{Current Obs Head} & \textbf{Input Seq} \\
    \midrule
    Cartpole             & $93\%$        & 2 LSTM, 1 Dense    & 1 LSTM, 1 Dense       & 1 Dense               & 50 \\
    Cartpole Seq         & $83\%$        & 2 LSTM, 1 Dense    & 2 LSTM, 1 Dense       & 1 Dense               & 50 \\
    Space Invader        & $86\%$        & 6 Conv, 3 LSTM, 2 Dense & 2 LSTM, 1 Dense  & 5 Conv, 2 Dense       & 10 \\
    Space Invader Seq    & $82\%$        & 6 Conv, 3 LSTM, 2 Dense & 2 LSTM, 1 Dense  & 5 Conv, 2 Dense       & 5 \\
    Pong                 & $97\%$        & 6 Conv, 3 LSTM, 2 Dense & 2 LSTM, 1 Dense  & 5 Conv, 2 Dense       & 2 \\
    Pong Seq             & $97\%$        & 6 Conv, 3 LSTM, 2 Dense & 2 LSTM, 1 Dense  & 5 Conv, 2 Dense       & 10 \\
    \midrule
    Average              & $90\%$        & -                   & -                    & - & -  \\
    \bottomrule
    \end{tabular}
    \label{tab:approximator}
\end{table*}

\subsection{Temporal Information Approximation}

As mentioned previously in~\Cref{sec:related}, the problem of approximating
a target agent is similar to an imitation learning problem. In this
paper we present a new reduction to a multi time-step learning problem.
The agent relies on a learned policy $\pi$ to act; its decision depends on its observation $s_t$, its previous actions $A_{t-1}$ and previous states $S_{t-1}$.
An approximation model predicts a sequence of future actions
instead of a single one.
We will explain how this opens up
temporal opportunities to the attackers in~\Cref{sec:temporal_attack}.
For convenience, we define the approximation model as:
\begin{equation}
    A^{f}_t = f(A_{t-1}, S_{t-1}, s_t),
\end{equation}
which output predicts a sequence of future actions
$(a_{t}, a_{t+1}, a_{t+2}, \cdots, a_{t+m})$.

Function $f$ is an approximator that takes sequence
inputs $A_{t-1}$ and $S_{t-1}$ together with the current state $s_t$.
We use $n$ and $m$ to represent input time steps and output time steps respectively;
this is equivalent to an input sequence having a length of $n$ and
an output sequence having a length of $m$.
The output sequence $A^{f}_{t}$ contains $m$ future actions.
Following the \bbox~setup, we only need to
observe how the agent is playing the game to build up a collection
of $((A_{t-1}, S_{t-1}, s_t), A^{f}_t)$ and hold this collection
as training data to train the approximator $f$.

We trained a sequence-to-sequence neural network as illustrated in~\Cref{fig:network} defined as $f$.
The network is multi-head to consume three different inputs $(A_{t-1}, S_{t-1}, s_t)$.
The seq2seq model is a multi-head one with heads focusing on $A_{t-1}$, $S_{t-1}$ and $s_t$ respectively.
Two of the inputs ($A_{t-1}$ and $S_{t-1}$) are sequence inputs and we used LSTMs in their relative paths to extract information, which is common in seq2seq learning \cite{hochreiter1997long,sutskever2014sequence}.
These two paths are shown as the blue and yellow blocks in \Cref{fig:network}.
The third path (red) contains input $s_t$ which represents the observation made at current state, so a single convolution followed by a fully connected layer is used to extract information.
We found that, depending on the game, the complexity of the head had to change to get a well-performing seq2seq model.
These heads affect the quality of information extraction, so we tuned them manually; a full per-game architecture for each head is shown in \Cref{tab:approximator}.
After information passes through each head, we duplicate the embedding $m$ times to form an output sequence with a temporal dimension of $m$.
These later blocks remain unchanged for different games, since empirically we found changing the heads is enough to provide us good accuracy on predicting actions on different games.
The later blocks include duplicating embeddings and aggregation of three different embeddings.
When duplication occurs, embeddings are duplicated $m$ times, where $m$ is the length of the output sequence.
In addition, these embeddings are aggregated by summation and fed into another fully-connected layer to produce a sequence of predicted actions.

Determining the number of input steps necessary for the input sequence is not trivial.
It is challenging to derive an optimal $n$ formally before training an approximator $f$, so we just have to search for it.
Luckily, the accuracy of a seq2seq model with various values of $n$ differs at an early stage of training.
In practice, we set a search budget of $N_t = 0.01N$, where $N$ is the number of training epochs.
\Cref{alg:train} describes the procedure of training a seq2seq model, where $[\,]$ denotes an empty list and the addition of two lists joins them sequentially.
First, we run a trained policy $\pi$ to produce action $a_t$; internally, $\pi$ might have stacked previous states or actions.
The sequence $E$ consists of the historical states and actions in an episode of game playing.
The set $D$ consists of multiple sequences $E$, and are used as data for training
($D_{\mathrm{train}}$) and evaluation ($D_{\mathrm{eval}}$).
Given an epoch count $N$, the selection of an input sequence length requires training of $n_{\mathrm{max}}$ seq2seq models, each trained for $N_t = 0.01 \times N$ epochs.
The $\mathrm{train}$ function requires an input sequence length, a training dataset and the number of epochs to execute training for the seq2seq model.
The $\mathrm{Split}$ function randomly shuffles the collected data, marking $90\%$ of the data as training data and $10\%$ as evaluation.
We found that the evaluation accuracy of the seq2seq model after training for a small number of epochs was enough to pick the optimal input sequence length ($n$).
After picking $n$, we then train with $N$ epochs to get our fully-trained seq2seq prediction model.
In our experiments, we used $N=500$ and $n_{\mathrm{max}}=50$.


\begin{figure*}[t]
	\centering
    \includegraphics[width=\linewidth]{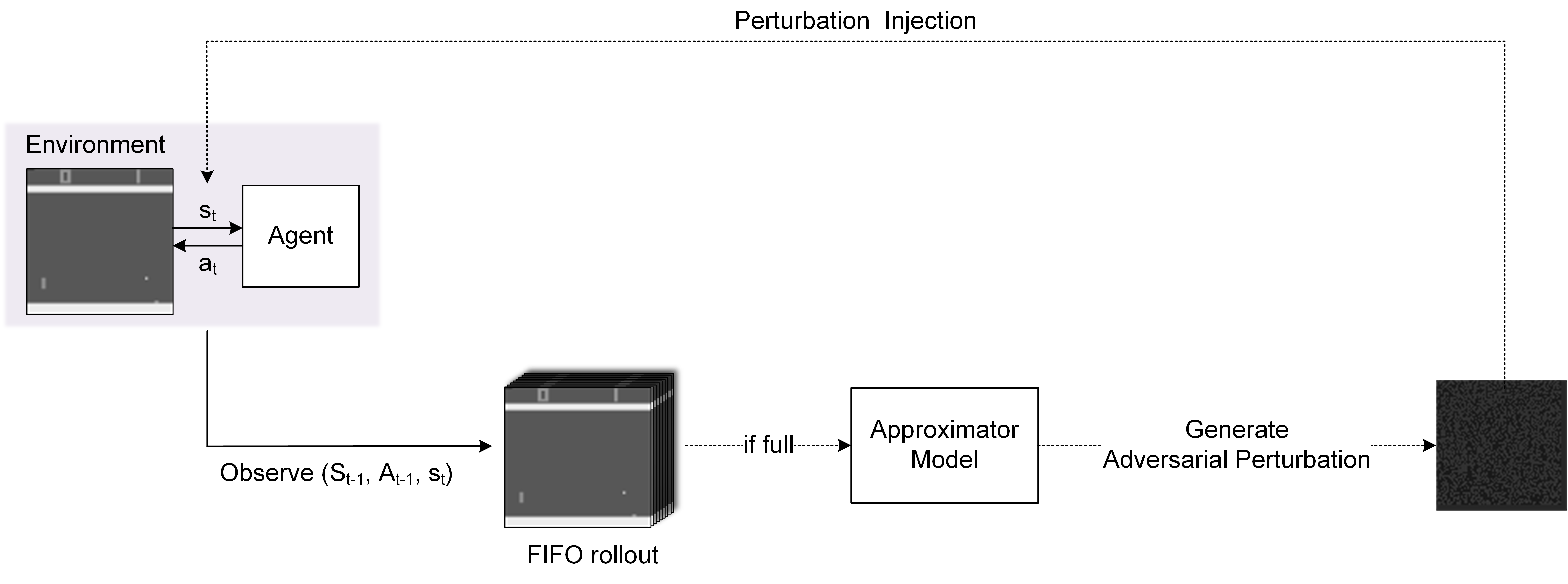}
    \caption{
        The attacker observes how a trained agent
        plays a game and can inject perturbations into the agent's observation $s_t$.
        We collect a sequence of observations to feed into the seq2seq model
        and generate adversarial perturbations to attack the trained agent.
        The adversarial perturbation in this figure is from an FGSM attack
        with $\epsilon=0.01$.
    }
    \label{fig:attack_flow}
\end{figure*}

\begin{algorithm}[!h]
\caption{Training Algorithm For Seq2Seq Model}
\label{alg:train}
\begin{algorithmic}[1]\small
\State{$D \gets \varnothing$}
\While{$|D| < N$}
    \State{$E \gets [\,]$}
    \State{$s_t = \mathrm{Init~Env}$}
    \While {($\mathrm{Game~not~done}$)}
        \State{$a_{t} = \pi(s_t)$}
        \State{$s_{t+1} = Env(a_{t})$}
        \State{$E \gets E + [s_t, a_{t}, s_{t+1}]$}
    \EndWhile{}
    \State{$D \gets D \cup E$}
\EndWhile
\State{$n \gets 0$}
\State{$\mathrm{acc}_m \gets 0$}
\State{$N_t \gets 0.01 N$}
\State{$D_{\mathrm{train}}, D_{\mathrm{eval}} \gets \mathrm{Split}(D)$}
\For {$n_i \in \{0, 1, ..., n_{\mathrm{max}}\}$}
    \State{$w \gets \mathrm{train}(n_i, D_{\mathrm{train}}, N_t)$}
    \State{$\mathrm{acc} \gets \mathrm{eval}(w, n_i, D_{\mathrm{eval}})$}
    \If {$\mathrm{acc} > \mathrm{acc}_m$}
        \State{$n \gets n_i$}
        \State{$\mathrm{acc}_m \gets \mathrm{acc}$}
    \EndIf{}
\EndFor{}
\State{$w \gets \mathrm{train}(n, D_{\mathrm{train}}, N)$}
\end{algorithmic}
\end{algorithm}

\subsection{Transferring Adversarial Samples}
Once we have a seq2seq model that approximates the target agent, we consider how to attack it.
Previous \bbox~attacks on convolutional neural networks can be mainly classified into two categories: gradient-based and decision-based.
Gradient-based attacks assume access to the gradients produced by the model. In the absence of the gradient access, the first step is to approximate the model in oracle fashion and then use the approximated gradient information. The transferability of the attack then depends on how well the model gets approximated.
Decision-based attacks take a different approach. They do not rely on the gradient information but try to find model's decision boundaries through continuous interaction with the model, to find a perturbation somewhere near the closest decision boundary~\cite{brendel2017decision}. In practice this takes a large number of queries.

In this paper, we use gradient-based attacks. The reasoning is two-fold. First, continuous querying is infeasible in RL. Second, the attack can still be used on the approximated agent, but in such a case we might as well use the gradients directly. They should work better and converge much faster.

In this paper, we have used a number of attacks of varying
complexity. As the weakest attacker we assumed Random Gaussian Noise -- an attack that does not actually use any information from our model.
It merely injects noise of a particular amplitude into the target agent's inputs, and serves as a baseline attack.


As a more sophisticated attack, we used FGSM~\cite{goodfellow2014explaining}.
The idea behind FGSM is to use the gradient information from a particular classifier to attack it.
As the most complex attack, we consider PGD~\cite{madry2018towards}, which can be seen as an iterative version of FGSM.
PGD is considered to be more complex as it requires iterative gradient computation per sample, but it produces
smaller adversarial perturbations.

Note that the list of attacks being tested in this paper is not exhaustive and many other complex attacks could be conducted.
CW~\cite{carlini2017towards} is generally considered as the strongest attack, and was used against RL agents by~\citeauthor{lin2017tactics}.
However, to get good misclassification rates and transferability, one needs to run CW for a very large number of iterations and with a large binary search factor.
Unlike adversarial machine learning in image classification where the evaluation can include only a thousand inputs, a single game episode in RL may contain thousands of decisions, making evaluation of CW on RL agents infeasible.
As we will highlight in later sections, a smaller perturbation size attack in RL in a \bbox~setup, does not necessarily translate to a direct decrease in agents' accumulated rewards.


The detailed attack flow is illustrated in~\Cref{fig:attack_flow}.
Our \bbox~attack starts after $n$ time steps when the rollout FIFO is full.
The rollout FIFO records playing histories so that inputs $(A_{t-1}, S_{t-1}, s_t)$ are prepared.
In our attack, the attacker has no access to the agents' internal states and parameters, or to the training environment.
After preparing the input sequence, the trained seq2seq model generates a prediction which can be used to craft adversarial perturbations.
The seq2seq model is multi-input, the adversarial perturbation is only generated for $s_t$, and the other two inputs $A_{t-1}$ and $S_{t-1}$ remain unchanged.
The perturbation is then injected to $s_t$ to attack the agent.
Although the seq2seq model requires an input sequence of length $n$ for its inputs $A_{t-1}$ and $S_{t-1}$, this does not mean we can only attack every $n$ steps.
We need to wait for the initial $n$ steps to start the first attack, but the following attacks can be generated consecutively since the FIFO is always full thereafter.

\begin{figure*}[!h]
  \minipage{0.22\textwidth}
    \includegraphics[width=0.7\linewidth]{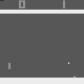}
  \endminipage\hfill
  \minipage{0.22\textwidth}
    \includegraphics[width=0.7\linewidth]{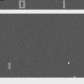}
  \endminipage\hfill
  \minipage{0.22\textwidth}%
    \includegraphics[width=0.7\linewidth]{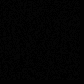}
  \endminipage\hfill
  \minipage{0.22\textwidth}%
    \includegraphics[width=0.7\linewidth]{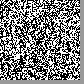}
  \endminipage
  \caption{
    Real adversarial inputs generated for the Pong game. The first image (leftmost)
    is the original input to an agent, the second image is the perturbed input.
    The third image shows the perturbation and the forth image is the
    same perturbation but rescaled to 0-255 for visibility,
    with 0 being black and 255 being white.
    If considered images been rescaled back to $0$ and $1$,
    the $l_2$ norm of this perturbation is $0.62$, and its $l_{\infty}$ norm is 0.01.
  }
  \label{fig:visual_sample}
  %
\end{figure*}

\section{Evaluation}
\label{sec:evaluation}
\subsection{Experiment Setup}
We trained agents using three different RL algorithms: DQN \cite{mnih2013playing}, A2C \cite{mnih2016asynchronous} and Rainbow \cite{hessel2018rainbow}.
The target games are Cartpole, Space Invader and Pong. For evaluation the game's randomness seed was reset for every episode.
Cartpole takes only $4$ input signals from the cart, while the other two are classic Atari games; we followed \citeauthor{mnih2013playing}'s method of cropping them to $84 \times 84$ image inputs~\cite{mnih2013playing}.
The RL algorithms and game setups are developed on Ray and RLLib \cite{moritz2018ray,liang2017rllib}.

As mentioned previously, we consider Random Gaussian Noise, FGSM \cite{goodfellow2014explaining} and PGD \cite{madry2018towards} attacks.
The implementation of the attacks was adapted from the commonly used adversarial ML library \textit{Cleverhans}~\cite{papernot2018cleverhans}.
Adaptations and extensions were made on the framework as its current form -- \textit{Cleverhans} -- does not support multi-input model or sequential outputs.

We control the sequence output $m$ to be $1$ and $10$ to illustrate the approximation possible for both short and long sequences of future actions.
The notion \textbf{Seq} is used for approximators that predicts $10$ sequential output actions.

\subsection{Seq2seq Approximation}
The seq2seq models vary in complexity when targeting various games and RL agents.
As mentioned previously, we alter the multi-head component in the seq2seq model as illustrated in~\Cref{fig:network} to adapt to various games.
The building blocks of each approximator are shown in \Cref{tab:approximator}.
The input sequence length of each approximator is determined using \Cref{alg:train}.
The approximators are then trained with Stochastic Gradient Descent (SGD) with a learning rate of $0.0001$.
We collect $N=500$ episodes of game play from trained agents and use this collected data as a training dataset.
The approximator, at every training time, takes bootstrapped training data from the $500$ episodes of playing experience with a batch size of $32$.
The seq2seq models are then tested on the unseen evaluation
data of the agent playing the game.
As shown in~\Cref{tab:approximator}, we achieve on average about $90\%$ accuracy on all seq2seq models.
We found Space Invaders a more challenging game to approximate using a seq2seq model. We hypothesise that this could be attributed to more complex agent interaction with the environment.
Another interesting observation is that Pong requires a small input sequence count, so the Pong agent effectively only requires short-term observations.
But whether predicting a single action or a sequence of $10$ actions in future time steps, our trained seq2seq model predicts them correctly with high accuracy.
The results in~\Cref{tab:approximator} shows that imitating an RL agent can be formed as a classification problem of predicting a future action sequence that the agent will perform given a history of the agent's action-observation pairs.

With the ability to predict the agent's future actions, we can construct adversarial samples using the approximation model and then transfer them into the real agent.
We visually demonstrate the adversarial perturbations in~\Cref{fig:visual_sample}.
The leftmost is the clean image and the second one is the adversarial image.
The generated adversarial input causes an action change for the underlying agent, but is visually the same as the original input.
In addition, we show the actual perturbation in the third picture and the rescaled ones in the fourth.
We rescaled the noise to $0-255$ range so that the locations of perturbations become clear.
In short, we summarise:
\begin{itemize}
  \item Multi-step passive imitation learning is possible, and our seq2seq
  model can predict future action sequence of RL agents with consistently
  high accuracy on the games and RL algorithms tested.
  \item Adversarial samples generated from the seq2seq model can be
  imperceptible by humans.
\end{itemize}

\begin{figure*}[!h]
  \minipage{0.33\textwidth}
    \includegraphics[width=\linewidth]{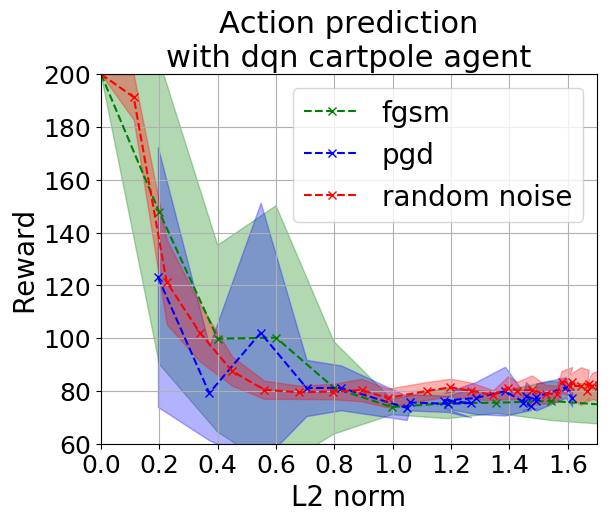}
  \endminipage\hfill
  \minipage{0.33\textwidth}
    \includegraphics[width=\linewidth]{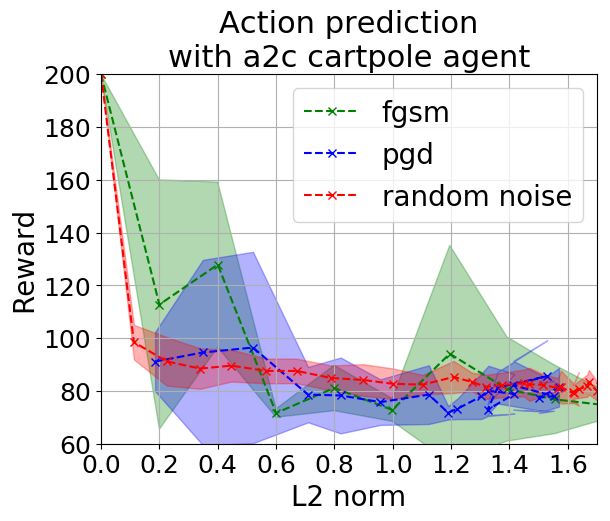}
  \endminipage\hfill
  \minipage{0.33\textwidth}%
    \includegraphics[width=\linewidth]{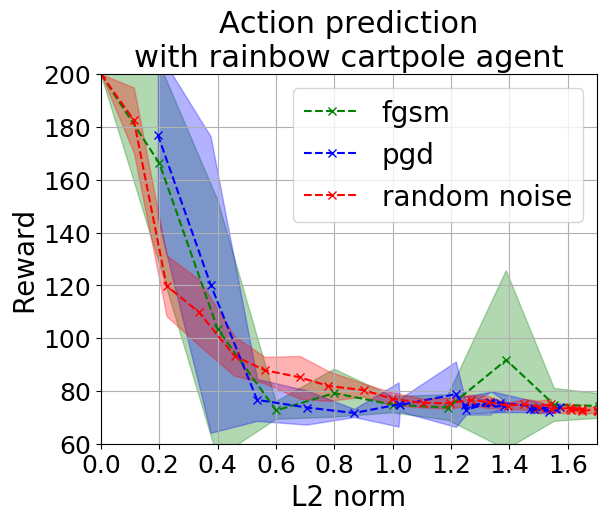}
  \endminipage
  \caption{
      \bbox~reward-focused attacks on DQN, A2C and Rainbow trained on Cartpole.
      The x-axis shows the size of $l_2$ norm of the adversarial samples, while
      the y-axis shows the reward of the attacked agents.
  }
  \label{fig:action_reward_attack_cartpole}
  %
\end{figure*}

\begin{figure*}[!h]
  \minipage{0.5\textwidth}
    \centering
    \includegraphics[width=0.6\linewidth]{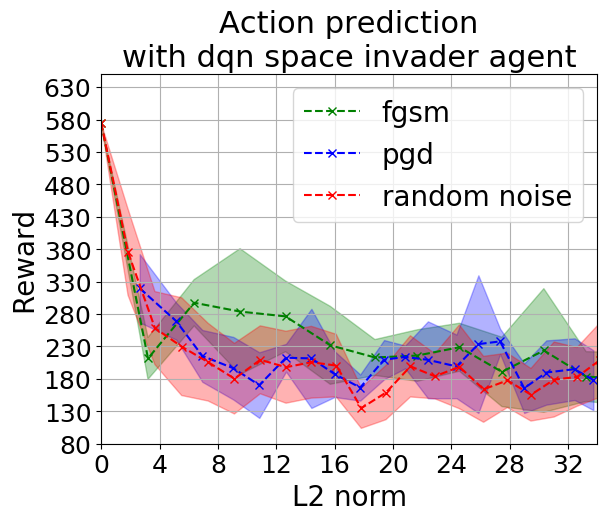}
  \endminipage\hfill
  \minipage{0.5\textwidth}%
    \centering
    \includegraphics[width=0.6\linewidth]{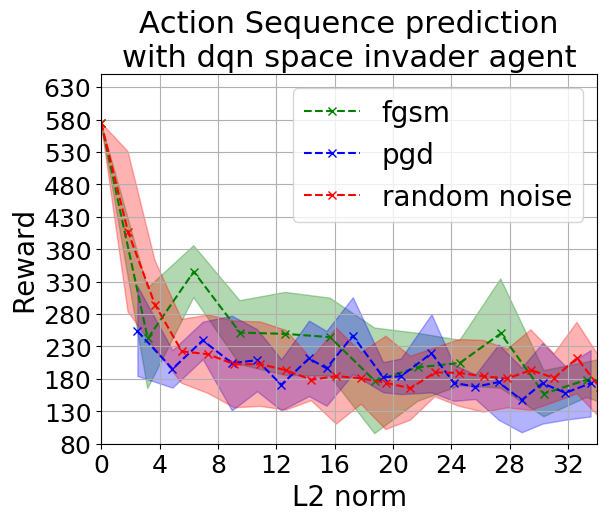}
  \endminipage
  \caption{
      \bbox~reward-focused attacks on DQN trained on Space Invader.
      The x-axis shows the size of $l_2$ norm of the adversarial samples, while
      the y-axis shows the reward of the attacked agents.
      `Action prediction' produces a single action but `Action Sequence' predicts 10 future actions.
  }
  \label{fig:action_reward_attack_space}
\end{figure*}

\begin{figure*}[!h]
  \minipage{0.5\textwidth}%
    \centering
    \includegraphics[width=0.7\linewidth]{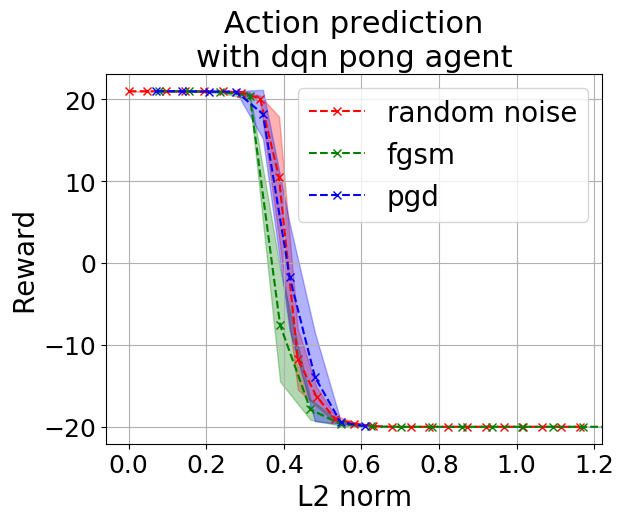}
  \endminipage
  \minipage{0.5\textwidth}%
    \centering
    \includegraphics[width=0.7\linewidth]{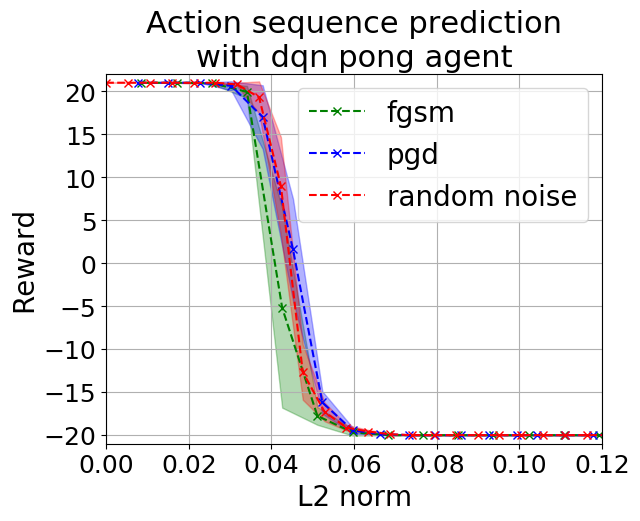}
  \endminipage
  \caption{
      \bbox~reward-focused attacks on DQN trained on Pong.
      The x-axis shows the size of $l_2$ norm of the adversarial samples, while
      the y-axis shows the reward of the attacked agents.
      `Action prediction' produces a single action but `Action Sequence' predicts 10 future actions.
  }
  \label{fig:action_reward_attack_pong}

  %
\end{figure*}
\begin{figure*}[!h]
    \minipage{0.33\textwidth}
      \includegraphics[width=\linewidth]{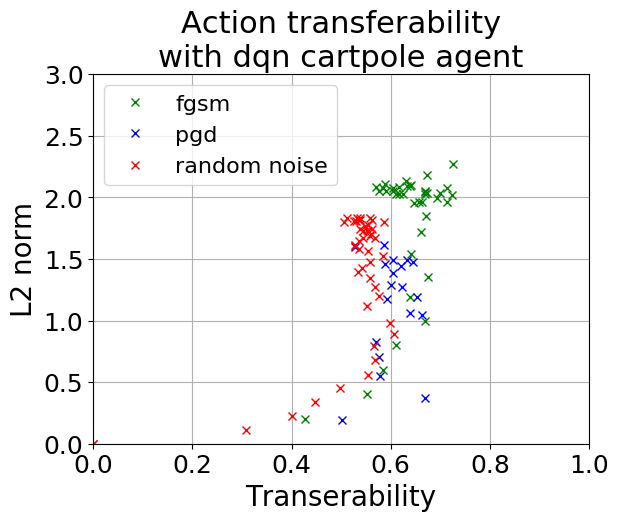}
    \endminipage\hfill
    \minipage{0.33\textwidth}
      \includegraphics[width=\linewidth]{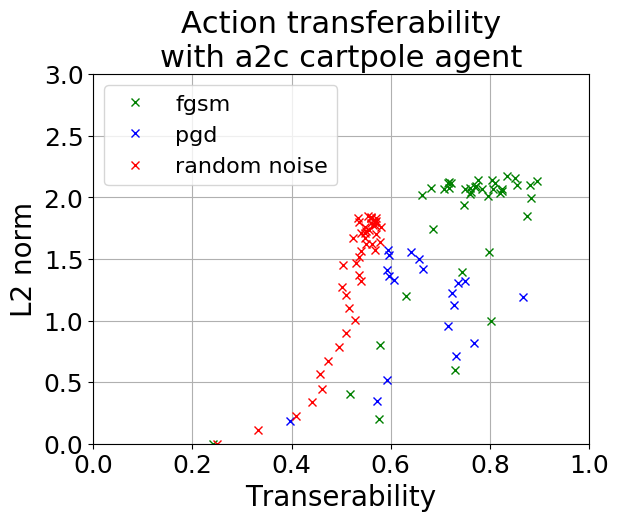}
    \endminipage\hfill
    \minipage{0.33\textwidth}%
      \includegraphics[width=\linewidth]{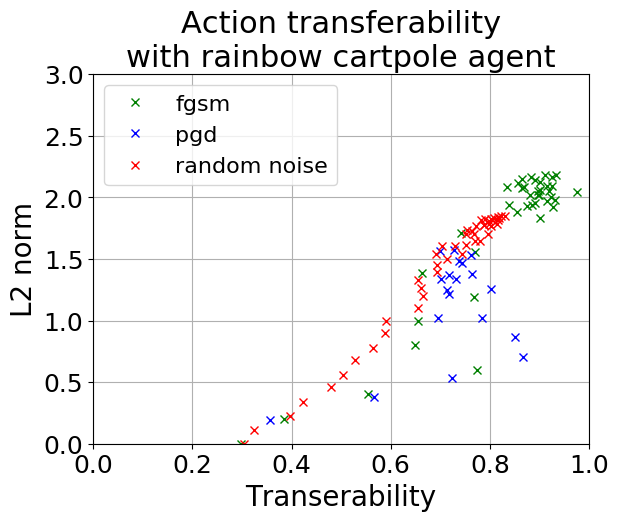}
    \endminipage
    \caption{
        Transferability of \bbox~reward-focused attacks on DQN, A2C and Rainbow trained on Cartpole.
        The x-axis shows the size of the transferability of
        the adversarial samples, namely the ratio between the number of samples that fooled the agents and the total number of samples generated
        from the seq2seq models.
        The y-axis presents the $l_2$ norm.
    }
    \label{fig:trans_action_reward_attack_cartpole}
    %
\end{figure*}

\subsection{Reward-focused Black-box Attack}
\label{sec:evaluate:reward}
Previous attacks in RL focus on reducing the accumulative reward of an agent in \wbox~or \gbox~setup \cite{lin2017tactics,huang2017adversarial}.
In this section, we discuss how our proposed \bbox~attack reduces an agent's accumulated reward.
The reward score is a direct measurement of the agent's game-playing quality; an efficient attack should be able to decrease it quickly by injecting perturbations.
\Cref{fig:action_reward_attack_cartpole} shows the performance of attacking agents trained for the Cartpole game.
The agents are trained with DQN, A2C and Rainbow.
Similarly, we show how agents trained with these three different RL algorithms perform on Atari games (Space Invader and Pong) in \Cref{fig:action_reward_attack_space} and \Cref{fig:action_reward_attack_pong}.
In \Cref{fig:action_reward_attack_cartpole}, we generate adversarial samples that aim at perturbing the agent's upcoming action $a_t$, so we call this the \textit{action prediction attack}.
The adversarial samples are generated using untargeted adversarial attacks, in other words, the attacker only wants to change the original action of the agent and does not ask it to change to any particular actions.
\Cref{fig:action_reward_attack_space} and \Cref{fig:action_reward_attack_pong} shows the attacks on Space Invader and Pong respectively.
In addition, we include the action sequence attacks, where the attacker focuses on flipping a random future action in the predicted sequence $a_t, a_{t+1} ... a_{t+m}$.
In these plots, we present error bars that are generated from $20$ distinct runs, and the mean values are extracted to plot the lines. Error bars represent a single standard deviation.
In our attack, we start to inject adversarial perturbations into the agent after collecting $n$ samples so that the FIFO rollout is full (\Cref{fig:attack_flow}).
However, the reward keeps accumulating in the initial collection stage, so even a powerful attack cannot reduce the score to zero.
Notice these attacks are untargeted, and the results generated are similar to \citeauthor{huang2017adversarial}'s, but \citeauthor{huang2017adversarial} ran their attacks in \wbox~and \gbox~setups~\cite{huang2017adversarial}.
The score decreases with increasing $l_2$ norms, meaning that agents are more vulnerable to larger perturbations.
From \Cref{fig:action_reward_attack_cartpole}, \Cref{fig:action_reward_attack_space} and \Cref{fig:action_reward_attack_pong}, we observe that all attacks have large variations.
This is because the performance of the attack depends heavily on the starting state $s_t$.
Intuitively, if we start to attack an agent when the agent is far away from any hazards it is harder to cause the agent to fail.
Next, we find that all attack methods show similar performance on the same game, but the effectiveness of attacks differ significantly across games.
For instance, a perturbation with an L2 norm of $0.8$ is sufficient to drive an agent to failure for Pong, but requires at least a L2 norm size of $4.0$ to affect Space Invaders.

We define transferability as the rate between the number of adversarial samples that caused misclassifications on the target agent side and the total number generated.
In~\Cref{fig:trans_action_reward_attack_cartpole}, we show a plot that measures transferability against the size of L2 norm across three different Rl algorithms for the Cartpole game.
In terms of transferability, FGSM and PGD outperform random noise.
As illustrated in~\Cref{fig:trans_action_reward_attack_cartpole}, adversarial attacks (FGSM and PGD) with the same $l_2$ norm have greater transferability in the Cartpole game across all RL algorithms.
In other words, adversarial inputs produced by FGSM and PGD are more likely to cause an agent to misbehave than random noise given the same perturbation budget.
But this does not translate to efficiency in reducing the rewards.
Many published works do not mention how reward decreases with input perturbation.
In our experiments, we found random noise generates comparable results to popular adversarial techniques in reward-based attacks given the same noise budget, and should have served as a baseline for work of this kind.
In particular, FGSM and PGD do not do significantly better than random jamming.

Finally, we want to note that we observe an effect similar in nature to the canonical imitation learning problem. As we start attacking the agent, the performance of the attacks starts dropping. We hypothesise that this behaviour is caused by the inability of the approximator to accurately capture states that he has not observed. 
For sequential decision making using imitation learning~\citeauthor{bagnell2015invitation} noted that learning errors cascade, resulting in learner encountering unknown states~\cite{bagnell2015invitation}. It seems to be also true for attacks based on the approximated models in a fully~\bbox~setup. 
\citeauthor{pomerleau1989alvinn} addressed this problem by learning policy to recover from mistakes~\cite{pomerleau1989alvinn}. That hints that, for \bbox~setup with multiple decisions, performance of the attacks is limited by how well you know the agent's mistake recovery policy.

This also has a detrimental effect on the attacks targeting specific actions of the agent e.g. the approach proposed by~\citeauthor{lin2017tactics}. It is unlikely that the attacker will be able to tell which of the non-preferred actions should be taken to attack the agent, if he has never observed the agent taking them.

For reward-focused \bbox~attacks, we summarise the following:
\begin{itemize}
  \item The attacks have large variations in terms of performance, the effectiveness of the attack is associated with where the attack starts.
  \item Various attack schemes have similar performance for the same game, although the agent playing the game might be trained with various RL algorithms. Nothing significantly outperforms random jamming.
  \item For attacks to be effective on different games, the required perturbation budgets vary significantly.
  \item FGSM and PGD show better transferability than random noise, meaning that samples generated from FGSM and PGD are more likely to flip the action of an agent.
  \item Performance of the attacks is bounded by the canonical imitation learning problem. As the agent gets approximated only on the actions that its prefers,
  it is hard to predict how its will react to the attacks in the non-preferred states.
\end{itemize}

\begin{figure*}[!ht]
    \minipage{0.5\textwidth}
      \centering
      \includegraphics[width=0.66\linewidth]{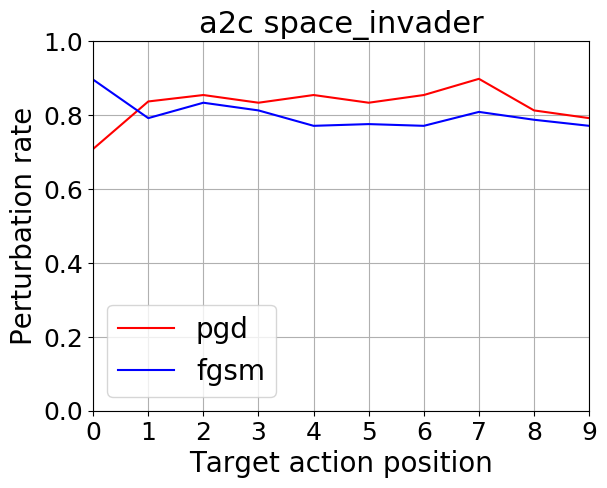}
    \endminipage\hfill
    \minipage{0.5\textwidth}
      \centering
      \includegraphics[width=0.66\linewidth]{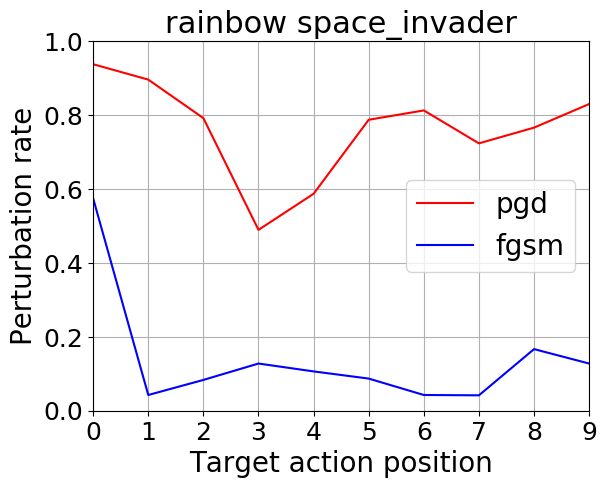}
    \endminipage\hfill
    \caption{
        \bbox~temporal-focused attacks on A2C and Rainbow trained on Space Invaders.
        The seq2seq model is trained against the DQN algorithm.
        The seq2seq model is then used to generate adversarial samples on A2C and Rainbow trained games to access transferability across RL algorithms.
        An adversarial image is sent only at $s_t$ but aims to perturb an action $n$ steps away $a_{t+n}$.
        The x-axis shows the targeted future step $n$, and the y-axis the perturbation rate averaged across $20$ runs.
    }
    \label{fig:attack_temporal_1}
\end{figure*}

\begin{figure*}[!ht]
    \minipage{0.5\textwidth}%
      \centering
      \includegraphics[width=0.66\linewidth]{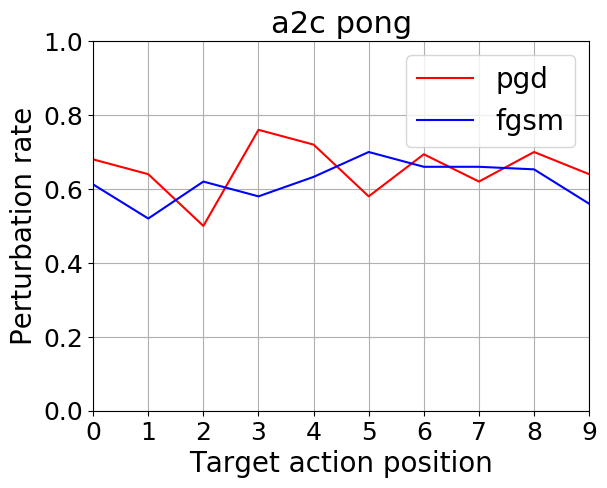}
    \endminipage
    \minipage{0.5\textwidth}%
      \centering
      \includegraphics[width=0.66\linewidth]{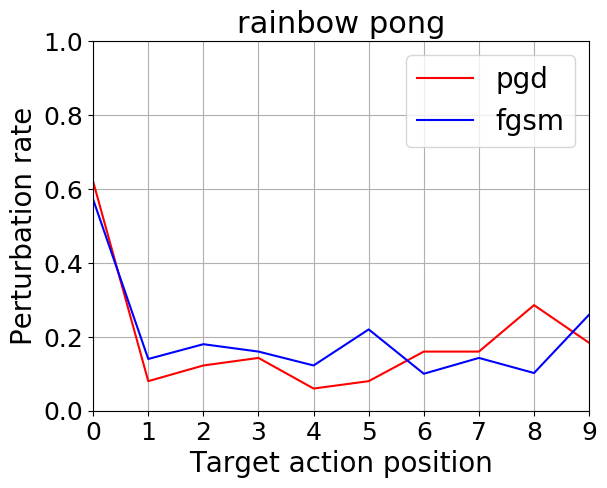}
    \endminipage
    \caption{
        \bbox~temporal-focused attacks on A2C and Rainbow trained on Pong.
        The seq2seq model is trained against the DQN algorithm.
        The seq2seq model is then used to generate adversarial samples on A2C and Rainbow trained games to access transferability across RL algorithms.
        An adversarial image is sent only at $s_t$ but aims to perturb an action $n$ steps away $a_{t+n}$.
        The x-axis shows the targeted future step $n$, and the y-axis the perturbation rate averaged across $20$ runs.
    }
    \label{fig:attack_temporal_2}
\end{figure*}

\subsection{Time-bomb Attack}

\label{sec:temporal_attack}

\begin{figure}[!ht]
  \centering
  \includegraphics[width=0.4\linewidth]{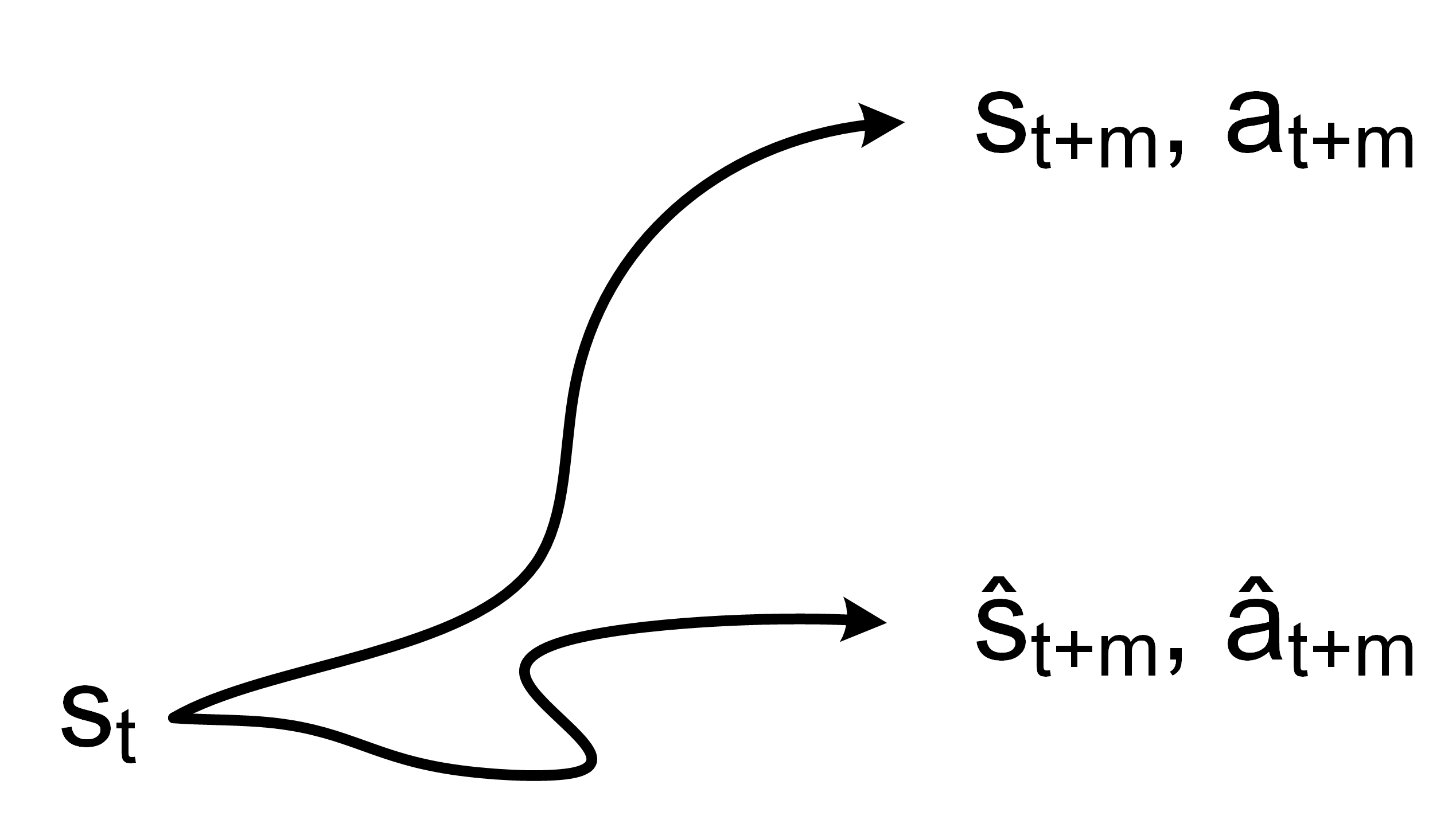}
  \caption{
    A high-level illustration of time-bomb attack.
    An adversarial input is injected at $s_t$, which
    causes the agent to misbehave from $s_{t+1}$ by following a different trajectory and finally triggers the adversarial time-bomb action $\hat a_{t+m}$ at $t+m$
  }
  \label{fig:time_bomb}
\end{figure}

Adversarial attacks do not do better than random noise jamming at reducing agent's rewards, but are efficient in action-targeted attacks. This section demonstrates a new temporal attack, which we called the time-bomb attack, and which uses the seq2seq model to produce an output sequence.

With its ability to forecast future actions, the \bbox~attack can now target a specific time delay.
Suppose the attacker's goal is not to decrease the game score of a deployed agent but to trigger adversarial actions after a delay.
For example, the attacker might want to cause an autonomous truck containing explosive fuel to crash five seconds in the future, to have time to get out of the way.
We illustrate from the POMDP perspective how such an attack would work in \Cref{fig:time_bomb}.
Consider an agent that has its original trajectory $s_t, ..., s_{t+m}$. We would like to change action $a_{t+m}$ to $\hat a_{t+m}$ with only a perturbation at $s_t$.
The perturbation at time $t$ will now cause the agent to follow a completely different trajectory so that an adversarial action $\hat a_{t+m}$ is possible at a state $\hat s_{t+m}$.
Surprisingly, the fact of asking the agent to follow a different trajectory often does little harm to the score, since the reward is an accumulative measurement.

We demonstrate the performance of the time-bomb attack in
\Cref{fig:attack_temporal_1,fig:attack_temporal_2}.
In both figures, the x-axis shows the time delay.
For instance, $1$ means we aim to perturb action $a_{t+1}$, and $x$ means $a_{t+x}$.
We only send adversarial noise at $s_t$, and all future observations made by the agents are clean.
The y-axis shows the success rate of such an action.
This temporal-focused attack shows that the success rate of perturbing a particular future action relates to the games and the agent's training algorithms.
We used DQN to train a seq2seq model and directly transfer the adversarial samples from the trained seq2seq model to both A2C and Rainbow agents.
The results in \Cref{fig:attack_temporal_1,fig:attack_temporal_2} show that this time-bomb attack works better on A2C trained agents than the Rainbow-trained agents, since Rainbow is a more complex algorithm.
In addition, the performance in Space Invaders is better than in Pong, which suggests that Pong is harder to sabotage in this way.
This makes intuitive sense since the game physics is much harder.
We limited the attacks in \Cref{fig:attack_temporal_1,fig:attack_temporal_2} to have $\epsilon=0.3$ which implies the $l_{\infty}$ norm is bounded by $0.3$.
This attack budget is picked to demonstrate the differences across games and RL algorithms.
With a larger attack budget (with $\epsilon$ bigger than $0.7$), we do observe all agents on all games getting attacked with success rates consistently higher than $70\%$.
We believe the time dimension of RL algorithms is not yet fully explored in the field of adversarial RL.
Our time-bomb attack shows that an attacker can in principle craft \bbox~attacks that perturb only the current time frame, but cause the target agent to misbehave after a specific delay.
Intuitively, this might not be too surprising, since there are many games where one player can set a trap for another.
The breakthrough here is to demonstrate it as a general attack on game-playing and similar RL agents.

The implication of time-bomb attack in RL is realistic and broad.
First, we only inject adversarial samples at a given time step, which proves that there are effective attacks on RL agents given only a short time window to attack.
In reality, constantly injecting adversarial noise into
the system can easily trigger detection~\cite{chen2019stateful,shumailov2018taboo,shumailov2019sitatapatra}.
Our \bbox~attack is thus more efficient in the temporal dimension.
Second, we show it is possible to intervene at the current time step but trigger a particular adversarial action at a future time step.
This type of attack can have broad implications, from spoiling an opponent's aim to fooling trading systems.
Finally, the described time-bomb attack is a general case of the path-planning attack shown by \citeauthor{lin2017tactics}~\cite{lin2017tactics}.

\section{Conclusion}

This paper offers three things: an improvement in the state of the art, a critique of the research methodology used thus far, and a new research challenge.

We explored how attackers can craft \bbox~attacks
against reinforcement learning (RL) agents. The \bbox~attack assumes that attackers have no access to any internal states or the training details of an RL agent. To the best of our knowledge, this is the first fully \bbox~attack against RL agents.
We discovered three things.

First, we can use seq2seq models to predict a sequence of future actions that an agent will perform, and use these to generate highly transferable adversarial samples. This improves the state of the art, as previous attacks were \wbox~or \gbox. We show that even in a \bbox~setup the attack will still work and disrupt the performance of an agent with an unknown objective function.

Second, although these adversarial samples are transferable,
they do not outperform random Gaussian noise as a means of reducing the game scores of trained RL agents. This
highlights a serious methodological deficiency in previous work on game-playing agents; random noise jamming should have been used as a baseline.

Our adversarial attacks do, however, have one advantage over random jamming: they can be used to trigger a trained agent to misbehave at a specific time in the future. This is our third discovery, and it appears to be a genuinely new type of attack; it potentially enables an attacker to use devices controlled by RL agents as time-bombs. Given that many games allow or even encourage players to set traps for other players, this might perhaps have been expected. But in applications other than games, the stakes can be higher. Where RL techniques are used to control safety-critical equipment, it will be necessary to find ways to ensure that such traps cannot be discovered and exploited by adversaries. How that might be done is the new research question posed by this paper.

\section*{Acknowledgements}
\textit{Partially supported with funds from Bosch-Forschungsstiftungim Stifterverband}

\bibliographystyle{IEEEtranN}
\bibliography{references}

\end{document}